\def\year{2022}\relax
\documentclass[letterpaper]{article} 

\usepackage{aaai22}  
\usepackage{times}  
\usepackage{helvet}  
\usepackage{courier}  
\usepackage[hyphens]{url}  
\usepackage{graphicx} 

\usepackage{natbib}  
\usepackage{caption} 
\DeclareCaptionStyle{ruled}{labelfont=normalfont,labelsep=colon,strut=off} 
\usepackage{algorithm}
\usepackage{algorithmic}
\usepackage{subfigure}
\usepackage{booktabs}
\usepackage{multirow}
\usepackage{color}

\usepackage{caption,tabularx,booktabs}

\usepackage{graphicx}

\usepackage{newfloat}
\usepackage{listings}
\usepackage{amsmath}
\floatstyle{ruled}
\newfloat{listing}{tb}{lst}{}
\floatname{listing}{Listing}
\title{Semantic-Based Few-Shot Learning by Interactive Psychometric Testing}
\author{
Lu Yin\thanks{Accepted at the AAAI-22
Workshop on Interactive Machine Learning (IML@AAAI’22)},
Vlado Menkovski,
Yulong Pei,
Mykola Pechenizkiy
}
\affiliations{
    Eindhoven University of Technology, Eindhoven 5600 MB, Netherlands\\
    \{l.yin,V.Menkovski,y.pei.1,m.pechenizkiy\}@tue.nl
}

\usepackage{bibentry}

\begin{document}

\maketitle

\begin{abstract}

Few-shot classification tasks aim to classify images in query sets based on only a few labeled examples in support sets.  Most studies usually assume that each image in a task has a single and unique class association. Under these assumptions, these algorithms may not be able to identify the proper class assignment when there is no exact matching between support and query classes.  For example, given a few images of lions, bikes, and apples to classify a tiger.  However, in a more general setting, we could consider the higher-level concept, the large carnivores, to match the tiger to the lion for semantic classification.  Existing studies rarely considered this situation due to the incompatibility of label-based supervision with complex conception relationships.  In this work,  we advance the few-shot learning towards this more challenging scenario, the semantic-based few-shot learning, and propose a method to address the paradigm by capturing the inner semantic relationships using interactive psychometric learning. The experiment results on the CIFAR-100 dataset show the superiority of our method for the semantic-based few-shot learning compared to the baseline.
\end{abstract}

\section{Introduction}

With enormous amounts of labeled data, deep learning methods have achieved impressive breakthroughs in various tasks. However, the need for large quantities of labeled samples is still a bottleneck in many real-world problems. For this reason, few-shot learning~\cite{lake2011one,vinyals2016matching} is proposed to emulate this by learning the transferable knowledge from the ``base" dataset where ample labeled samples are available to generalize to another ``novel" dataset which has very few labeled training
examples. A popular approach for this problem is meta-learning based phase~\cite{snell2017prototypical,finn2017model} which follows the episodic training procedure to mimic the few-shot tasks.  In each few-shot task, a few labeled examples (the support set) are given to predict classes for the unlabeled samples (the query set).

\begin{figure}[tbp]
\centering
    \subfigure{Typical four-way one-shot learning task}{
        \includegraphics[width=0.45\textwidth]{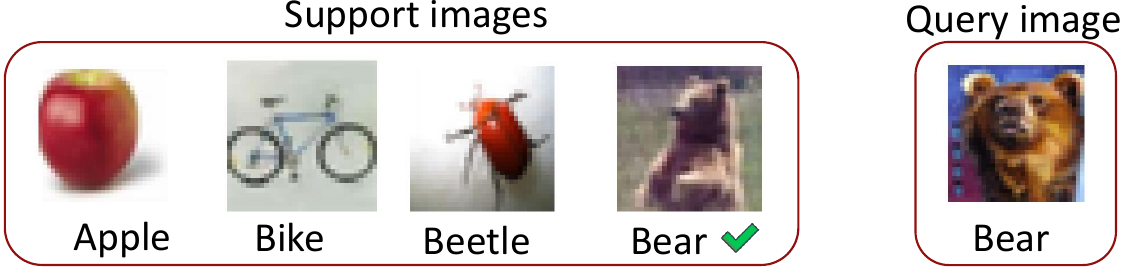}\label{FSL}
    }
    
    \subfigure{Semantic-based four-way one-shot learning task}{
        \includegraphics[width=0.45\textwidth]{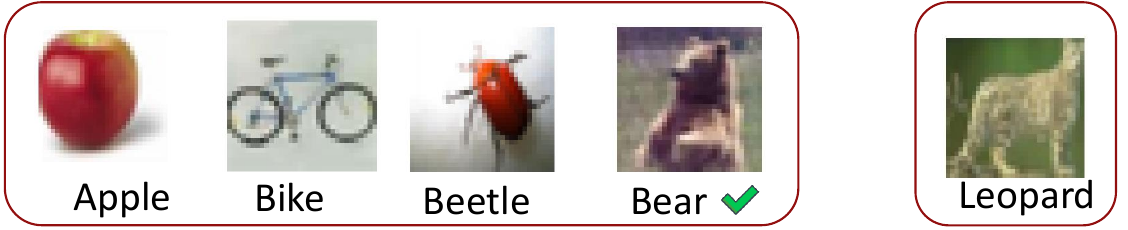}\label{SFSL1}
    }
    
    \subfigure{Semantic-based four-way one-shot learning task}{
        \includegraphics[width=0.45\textwidth]{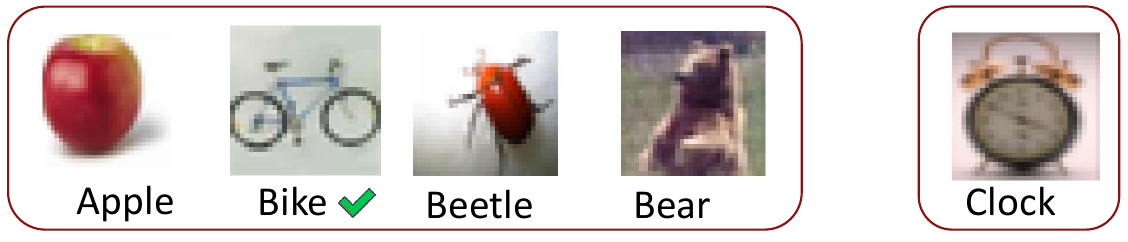}\label{SFSL2}
    }

\caption{ Different settings of few-shot learning tasks. The first row follows a typical four-way one-shot learning setting. The class of the query image matches one of the support set labels. In the second task, the typical few-shot learning model might fail to identify the query image when there is no exact class matching. However, if we could consider the higher-level semantic concept of the carnivores, a correct assignment could still be made by matching bear to leopard. A similar prediction could also be made if we consider the concept of non-living things in the last task.  }
    \label{FSl_compare}
\vspace{-0.5cm}
\end{figure}

While these formulations have made significant progress, the underlying assumption is that each data point from the support set and query set has a single and uniquely identified class association, and the query image must precisely match one of the support set classes.  However,  as illustrated in the last two rows in Figure~\ref{FSl_compare}, the few-shot learning models that are capable of dealing with classification based on the predefined classes may not be able to identify the right class assignment when there is no exact class matching.  

In a more general setting, if considering the concept at a higher level, e.g., whether they are large carnivores or living things in Figure~\ref{FSl_compare}, one could determine the right class association.  Humans are very capable of inferring these concepts on a higher level, while typical few-shot learning algorithms are not specifically designed for this under single discriminating class descriptions. They treat each class equally without considering their intra hierarchical semantic relationships.   One possible reason for this limitation might be  the supervision approach: the traditional label-based supervision is incompatible with the complex conception hierarchy.  Fortunately, much progress has been made in learning from other types of supervision, such as psychometric testing~\cite{gescheider2013psychophysics}.  While label-based supervision reduces the comprehensive semantic relationships to given discrete labels, these psychometric testing based methods could elicit the relative conception similarities and full-depth of knowledge by transmitting the annotations progress to pair or triplet comparisons. Then the elicited knowledge could be used for other downstream tasks such as clustering or segmentation~\cite{yin2021hierarchical,yin2020knowledge}.  Enabled with such techniques, our work aims to extend the capabilities of few-shot learning models towards a more challenging setting, the semantic-based few-shot learning.

To be specific, we assume there is a shared concept hierarchy covering both base and novel classes. Self-supervised learning (SSL) is applied for feature learning at the first stage. The interactive psychometric testing is then followed to capture the similarities of the semantic concepts from base dataset.  We use these semantic similarities to fine-tune the learned features from SSL, and map them to a semantic embedding space where we transfer the learned hierarchical knowledge from base classes to novel classes for semantic few-shot prediction. 

Our contributions could be summarized as follows.
\begin{itemize}
\item[*]  We define a new problem setting, the semantic-based few-shot learning. It aims to identify the correct assignment to query image by higher-level concepts when there is no class matching between query and support images.
\item[*]  We analyze the limitations of label-based supervision under the semantic-based few-shot learning setting and propose a psychometric learning based approach to tackle this problem.
\item[*] We evaluate our method by comparing it with a typical few-shot baseline (prototype network~\cite{snell2017prototypical}) on CIFAR-100 dataset~\cite{krizhevsky2009learning}. The results demonstrate that our method could significantly outperform this baseline in semantic-based few-shot learning even using fewer annotations from base data.

\end{itemize}

\section{Related Work}

There are three lines of research closely related to our work:  psychometric testing, few-shot learning, and self-supervised learning.

\noindent\textbf{Psychometric testing} Psychometric testing~\cite{gescheider2013psychophysics} aims to study the perceptual processes under measurable psychical stimuli such as tones with different intensity or lights with various brightness.
In general, two types of psychometric experiments could be carried. Firstly, the absolute threshold based method tries to detect the point of stimulus intensity that could be noticed by a participant. For example, how many hairs are touched to the back of hand before a participant could notice. Secondly, the discriminative based experiments aim to find the slightest difference  between two stimuli that a participant could perceive. Participants might be asked to describe the difference in direction or magnitude between these two stimuli or forced to choose between the stimuli concerning a specific parameter of interest (also known as two-alternative-force choice (2AFC) test~\cite{fechner1860elemente}). Some scholars extend the 2AFC to M-AFC methods~\cite{decarlo2012signal} by comparing $M$ stimuli in one test to elicit the subjects' perception of more complex multimedia such as videos or images ~\cite{son2006x,feng2014methodology,yin2021hierarchical,yin2020knowledge}. 
In our work, we take advantage of the 3-AFC method to align with our loss function. Three samples are presented in one test to elicit the annotator's perception regarding the conception similarity.


\noindent\textbf{Few-Shot Learning} Meta-learning (learning to learn) has gained increasing attention in the machine learning community, and one of its well-known applications is few-shot learning. Three main approaches have emerged to solve this problem. Metric learning based methods aim to learn a shared metric in feature space for few-shot prediction, such as prototypical network~\cite{snell2017prototypical}, relation networks~\cite{sung2018learning} and matching networks~\cite{vinyals2016matching}. 
Optimization based methods follow the idea of modifying the gradient-based optimization to adapt to novel tasks~\cite{nichol2018reptile,finn2017model,gidaris2018dynamic}. Memory based approaches~\cite{finn2017model,he2020memory} adopt extra memory components for novel concepts learning, and new samples could be compared to historical information in the memories. 

While these frameworks lead to significant progress, little attention has been paid to leveraging the knowledge hierarchy and dealing with the situation when there is no precise label matching between query images and support images, i.e., the semantic-based few-shot learning scenario.

\noindent\textbf{Self-Supervised Learning} When human supervision is expensive to obtain, self-supervised learning could be a general framework to learn features without human annotations by solving pretext tasks.  Various pretexts have been studied for learning useful image representation. For example,   predicting missing parts of the input image~\cite{trinh2019selfie,larsson2016learning,pathak2016context,zhang2016colorful,zhang2017split}, the image angle under rotation transformation~\cite{gidaris2018unsupervised}, the patch location, or the number of objects~\cite{noroozi2017representation}. Recently, another line of researches follows the paradigm of contrastive learning~\cite{bachman2019learning,chen2020simple,he2020momentum,henaff2020data,hjelm2018learning,misra2020self,oord2018representation,wu2018unsupervised} and get the state of the art performance. The learned image features could be utilized for downstream tasks such as image retrieval or fine-tuning for classification. In our work, we take advantage of the SimCLR~\cite{chen2020simple} framework and fine-tune the learned features with psychometric testing for semantic image representations.

\section{Semantic-Based Few-Shot Learning}

\begin{figure}[htbp]
\centering

    \subfigure{
        \includegraphics[width=0.45\textwidth]{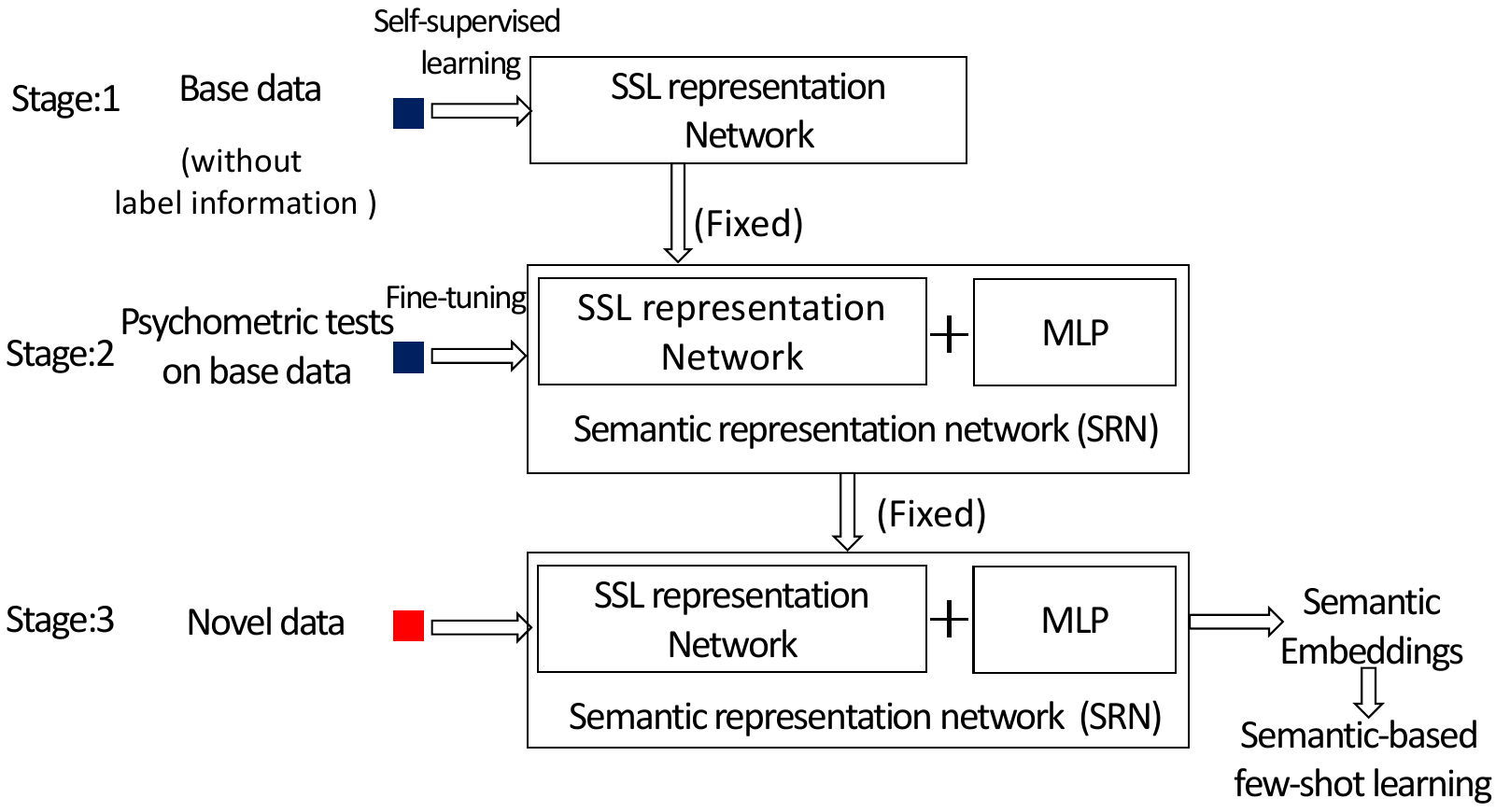}
    }

\caption{Overview of the proposed method.}
    \label{system}
\end{figure}

Our proposed framework contains three parts. First, as we aim to tackle the limitation caused by label-based supervision,  we assume no label information is provided in advance. Self-supervised learning (SSL) is applied for representation learning in the first stage. Next, we adopt a psychometric testing procedure~\cite{gescheider2013psychophysics} that relies on discriminative testing to obtain transferable semantic conception relationships.  The elicited conception similarities are then used to fine-tune the features learned by SSL using  a multi-layer perceptron (MLP)~\cite{friedman2017elements} in a semantic representation network. In the last stage,  with the fine-tuned network, we could search for each query's most semantically similar image in support set by Euclidean distances, even when the target and query images are not sharing the same class. We illustrate our whole framework in Figure~\ref{system}.

\vspace{-0.5cm}




\subsection{Problem Formulation}

Consider the situation we are given a base dataset contains classes $C_{base}$ with adequate  labeled images, and a novel dataset contains classes $C_{novel}$ where only a few labeled samples are available per class. There is no overlapping with these two datasets, i.e., $C_{base}{\cap}C_{novel}={\emptyset}$.  The general idea of the few-shot problem is taking advantage of the sufficient labeled samples in $C_{base}$ to obtain a good classifier for the novel class  $C_{novel}$.  In a standard $N$-way $K$-shot classification task, we random sample $N$ classes from novel class $C_{novel}$ with $K$ samples per class to form the support set, and sample query images from the same $N$ classes to create the query set. We aim to classify the query images into these $N$ classes based on the support set.

\begin{figure}[htbp]
\centering
    \subfigure{
        \includegraphics[width=0.4\textwidth]{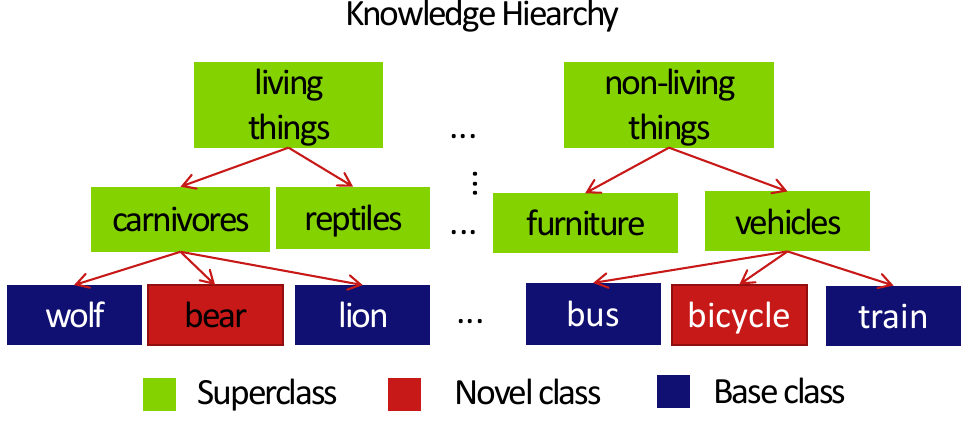}
    }

\caption{CIFAR-100 with knowledge hierarchy.}
    \label{Cifar_100}
\end{figure}

Then we extend the problem to the semantic-based few-shot learning scenario. Assume we have a conception tree $ G = (V, E)$ where $V$ means the nodes and $E$ are edges.  The bottom  layer class $C = {c_1, . . . , c_n} {\in} V$ denotes the lowest level of concepts that we concern, and could merge to more general concepts (superclass nodes) if they are conceptually similar. An example for such a structure is given in Figure~\ref{Cifar_100}. The base class  $C_{base}$ and novel class $C_{novel}$ are represented as the leaf nodes and share the same superclasses nodes. As we aim to solve this problem without label-based supervision, we are not able to specify a few-shot task using the label information as the typical few-shot learning setting, i.e., sampling multiple images with the same labels to create a class in support set.  Therefore, we random sample $N$ image without specifying their classes from the  $C_{novel}$ to build the support set and sample one image as a query to form a \textbf{\textit{N}-way \textit{1}-shot} semantic-based few-shot learning task. Our goal is to find the most semantically similar image from support set to a query.

The semantic distance between two samples $(x,y)$  is defined by the height of the lowest common subsumer (LCS) of these samples divided by the height of the hierarchy~\cite{barz2019hierarchy,verma2012learning}:

\begin{equation}
D_s(x,y)=\frac{height(lcs(x,y))}{max_{w{\in}V}height(w)} 
\end{equation}

As $D_s(x,y)$  ranges from $0$ to $1$, we could define the semantic similarity by:
\begin{equation}
S_s(x,y)=1-D_s(x,y)
\label{equ_SS}
\end{equation}

An example could be seen in Figure~\ref{Cifar_100}. The LCS of $wolf$ and $lion$ are $carnivores$, and the height of the hierarchical tree is 3. Therefore $D_s(wolf,lion) = \frac{1}{3}$, and $S_s(wolf,lion) = \frac{2}{3}$. Note that the typical few-shot learning is a special case when $S_s(x,y)=1$, in which $x$ is the query image, $y$ is from support set, and $x,y$ belong to a same leaf node.

\subsection{Self-Supervised Feature Learning}
We use self-supervised learning to learn the image features from $C_{base}$ before using psychometric testing for fine-tuning.  SimCLR~\cite{chen2020simple} framework is applied in our work for its conciseness and good performance. It learns representation by maximizing the similarity between two views (augmentations) of the same image. 

 From $C_{base}$, we randomly sample $N$ images each batch and create two random augmentation views for each image to form  $2N$ data points.  Each data pair generated from the same image is considered a positive pair, or a negative pair if it's from different images. The contrastive loss function for a mini-batch could be written as:
\begin{equation}
L_{self}=-\sum_{i=1}^{2N}log\frac{exp(sim(z_i,z_{j(i)}/\tau))}{\sum_{a{\in}A(i)} exp(sim(z_i,z_a)/\tau)}
\end{equation}
where $z_i=g(f(x_i)$),  $f({\cdot})$ a neural network called encoder to extract features from augmented images,  $g({\cdot})$is the projection head that maps features to a space where contrastive loss is applied. Cosine similarity $sim(u, v)$ is adopted to measure the similarity of $u$ and $v$  by the dot product between their $L_2$ normalized features.   $\tau$ denotes a scalar temperature parameter.  $i$ is the index or all the $2N$  augmented views of images.   $ j(i)$ is the index of positive view to image $i$ and  $A(i)$ is the set of all indices except $i$.

\subsection{Psychometric Testing}

Different from label-based supervision, we apply three-alternative-force choice (3AFC)~\cite{decarlo2012signal} psychometric tests to elicit the semantic perceptions from $C_{base}$. These perceptions could be transferred to $C_{novel}$ through the shared high-level conceptions (superclasses) in the  hierarchical knowledge tree (as shown in Figure~\ref{Cifar_100}).

\begin{figure}[htbp]
\centering
\includegraphics[width=0.45\textwidth]{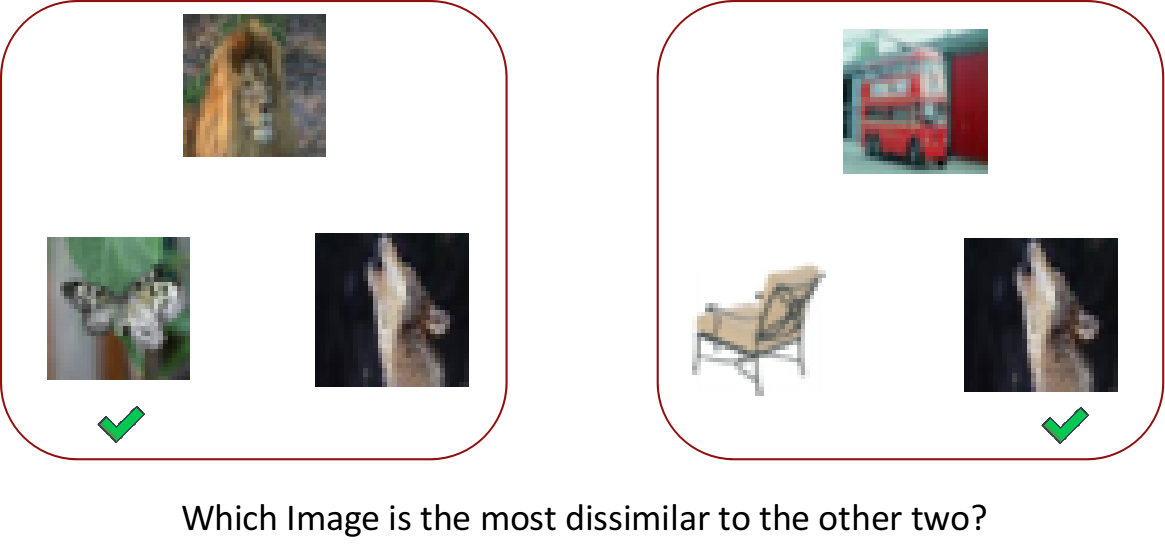} 
\caption{Examples of two 3AFC psychometric testings. In the first test, annotators tend to choose the butterfly as the most dissimilar one since other two are large carnivores. In the second test, annotators are more likely to choose the wolf because it is the only living things. }
\label{3afc}
\end{figure}

To be specific, we sample three images from $C_{base}$ and ask the annotators to choose the most dissimilar one (see Figure~\ref{3afc}). By carrying this simple task, perceptions of conception similarities are obtained.

Next, a semantic representation network (SRN) is built to map these perceived conception similarities to embedding distances. Specifically, we add a multi-layer perceptron (MLP) with a single hidden layer on top of the representations learned from SSL, freeze the SSL network, and fine-tune the MLP by the following dual-triplet loss  function~\cite{yin2020knowledge,yin2021hierarchical} :

\begin{equation}
\begin{aligned}
L=\sum_{i=1}^N \left[d(x_{p1}^i,x_{p2}^i) - d(x_{n}^i,x_{p1}^i) ,+m \right]_+ \\
+ \left[ d(x_{p1}^i,x_{p2}^i) - d(x_{n}^i,x_{p2}^i) ,+m \right]_+  \label{n_loss}   
\end{aligned}
\end{equation}where $x_{n}$ is the negative image chosen by annotator, $x_{p1},x_{p2}$ are two unpicked positive images that have closer concept similarity  at the 3AFC tests (see Figure~\ref{3afc}). 
 $d(x,y)$ denotes the these two points' Euclidean distance between the normalized features extracted by our semantic representation network. $N$ is the number of psychometric tests in a mini-batch.

This loss function encourages images that the annotator perceives similar to be close to each other and enforces a distance margin  $m$  between positive pairs and  negative pairs.

\subsection{Semantic-Based Few-Shot Prediction}

After fine-tuning our proposed network with  3-AFC tests from $C_{base}$,  we could extract visual features for image samples from $C_{novel}$ using this network and apply the nearest neighbor search method for semantic-based few-shot learning prediction. Specifically, for a query image in a task, we compute its normalized Euclidean distance to each support sample and find the nearest one, which is the predicted most semantically similar image to the query when considering a higher-level concept.

\section{Experiments and Discussion}

\begin{table*}[htbp]
\newcommand{\tabincell}[2]{\begin{tabular}{@{}#1@{}}#2\end{tabular}}  
\centering
\caption{Comparison with the baseline.}
\scalebox{0.93} {
\begin{tabular}{|c|c|c|c|c|c|c|}  
\hline
\multirow{2}{*}{Model} & 
\multirow{2}{*}{\tabincell{c}{Annotation\\  Type} }& 
\multirow{2}{*}{\tabincell{c}{Number of   \\  Annotations {($C_{base}$})}}  & 
\multicolumn{2}{c|} {5-way 1-shot Acc(\%)} & 
\multicolumn{2}{c|} {20-way 1-shot Acc(\%)} \\
\cline{4-7}
    &   &  & Typical & Semantic& Typical & Semantic\\ 
\hline
PN~\cite{snell2017prototypical} & Label Based & 36000 & \textbf{57.52}& 42.37& \textbf{31.18}& 19.81\\  
SRN(Ours)  & \tabincell{c}{Psychometric\\ Testing}&\textbf{1000}& 52.57 & \textbf{52.35} & 28.75& \textbf{27.16}\\  
\hline
\end{tabular}}
\label{Compare}
\end{table*}

Since the label-based supervision is a bottleneck that limits the models' potential in the semantic-based few-shot learning setting, we assume no label information and no conception structure are preprovided for both $C_{base}$ and $C_{novel}$. However, we are then not able to assess whether the semantic assignment to query is correct using the defined semantic similarity metric (see Equation~\ref{equ_SS}). Therefore, we simulate a virtual annotator who always precisely responds to the 3AFC tests based on a  given knowledge hierarchy, so that the accuracy could be measured in an objective manner by this semantic similarity.

We evaluate our model on CIFAR-100 dataset under three metrics: the typical few-shot learning accuracy, the semantic-based few-shot learning accuracy, and the required annotation numbers. Then we investigate how the number of psychometric test responses impacts the model's performance. Besides, a TSNE visualization~\cite{van2008visualizing} of the learned features is plotted for an intuitive understanding.

\noindent\textbf{Dataset} We use the CIFAR-100 in our experiment and build an inner conception hierarchy tree based on the preprovided coarse and finer labels.  Besides,  we build another layer on top of the coarse level labels by distinguishing living from non-living things. A three-layer conception tree is then created, which includes 2, 10, 100 nodes from top to bottom layers, as illustrated in Figure~\ref{Cifar_100}.  60 classes are randomly sampled from the bottom layer as base classes, and the rest 40 classes are used for novel classes.

\noindent\textbf{Few-shot learning accuracy} Note when there is a label matching between the query image and support images, i.e., the semantic similarity is equal to 1, the semantic-based few-shot learning problem is then transmitted to a typical few-shot learning problem. We choose the  prototypical network~\cite{snell2017prototypical} as a baseline and compared it with our proposed method in both typical few-shot learning accuracy and semantic-based few-shot learning accuracy.

In our work, we use the SGD optimizer with momentum 0.9, and set the decay factor to 0.1. When extracting image features in SSL, ResNet50~\cite{he2016deep} is applied as backbone and are trained for 1000 epochs with 128 batch-size. The learning rate decays from 0.5 at epoch 700, 800 and 900. When fine-tuning by psychometric responses, margin value, learning rate,  training epochs are set to 0.4, 0.001, 15 respectively.  During prototypical network training,  we use the same backbone of SSL for a fair comparison, train the model 100 epochs with 10000 tasks each epoch, and set the learning rate to 0.1 that decays every 20 epochs.

The results are reported in Table~\ref{Compare}. It could be seen without losing too much accuracy of typical one-shot learning (decreasing by $4.95\%$ in 5-way, and  $2.43\%$ in 20-way).
We could boost the ability of semantic-based one-shot learning significantly (increasing by $9.98\%$ in 5-way, and  $7.35\%$ in 20-way). Furthermore, the annotation burdens on base data are dramatically released from 36000 times label-based annotations to 1000 times psychometric testings.

\begin{figure}[htbp]
\centering
    \subfigure{}{
        \includegraphics[width=0.35\textwidth]{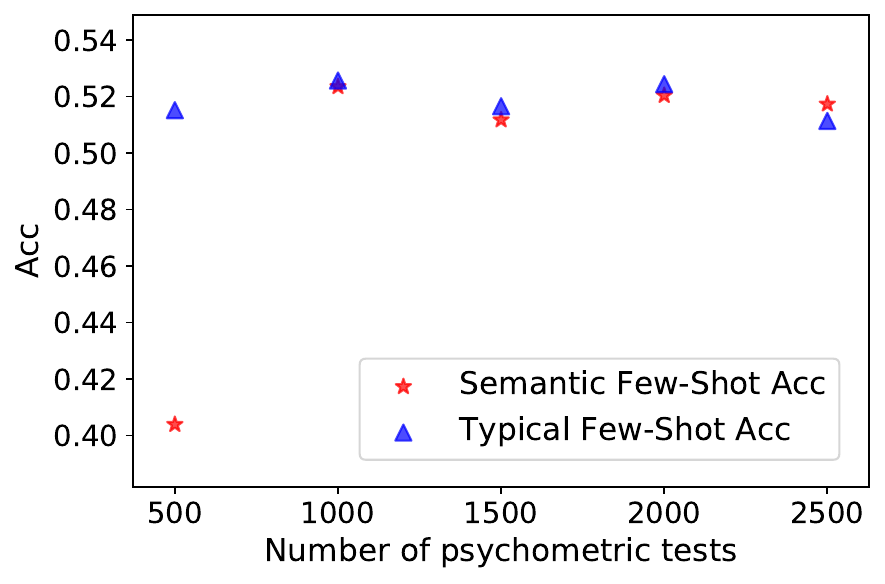}
    }

\caption{5-way 1-shot learning accuracy under different number of psychometric tests.}
    \label{semantic_acc}
\end{figure}

\noindent\textbf{Impact of the number of psychometric test responses}  We 
train our model using 500 psychometric tests in the first iteration and add 500 more tests to retrain the model in each of the following iterations.  The model is evaluated under the 5-way 1-shot scenario and we plot the results in Figure~\ref{semantic_acc}.  It could be noticed that the accuracy of typical few-shot learning remains steady with different numbers of psychometric tests. That is because our psychometric tests only aim to provide semantic constrain rather than learning discriminative features.  We also find that the ability of semantic few-shot learning gets a noticeable improvement when increasing training samples from 500 to 1000 tests but keeps stuck after that. The possible reason might be that with the help of pre-trained SSL features, we could easily get a high accuracy using only a few psychometric tests. However,  as we only fine-tune on MLP without training the whole network, the semantic few-shot accuracy would quickly reach a bottleneck even with more psychometric responses.

\vspace{-0.2cm}

\begin{figure}[htbp]
\centering
    \subfigure{}{
        \includegraphics[width=0.2\textwidth]{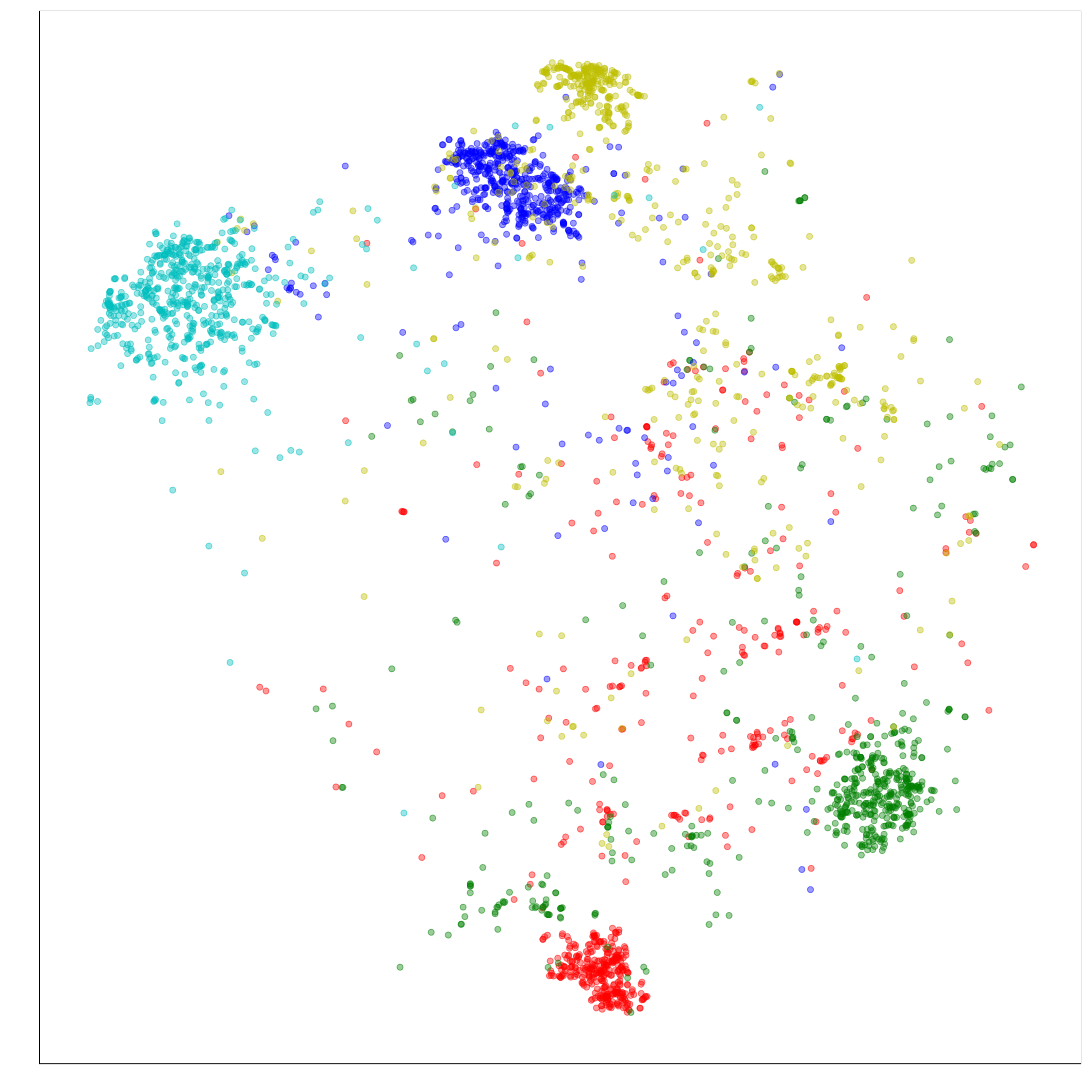}}
    \subfigure{}{
        \includegraphics[width=0.2\textwidth]{ 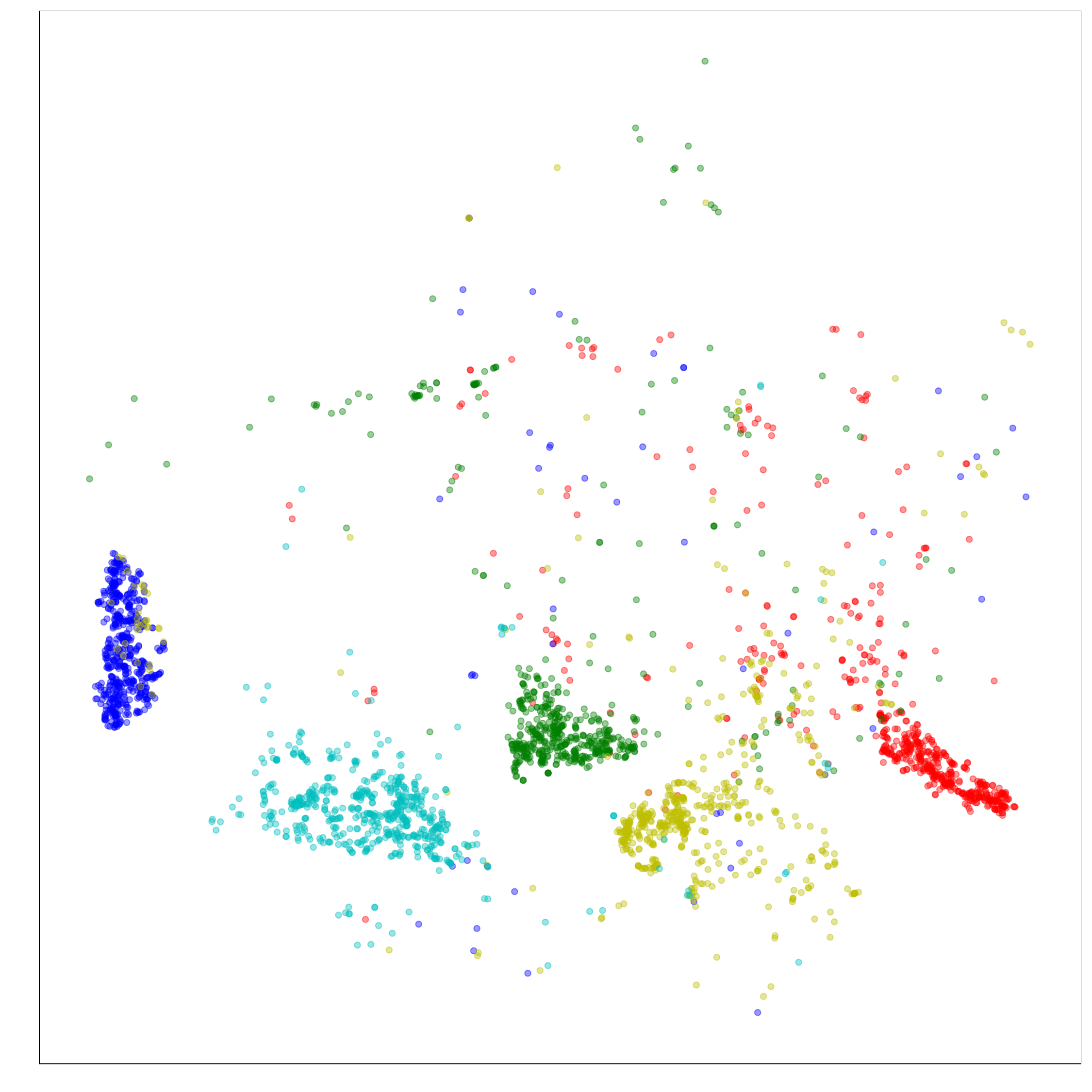} }
    
    \subfigure{}{
        \includegraphics[width=0.45\textwidth]{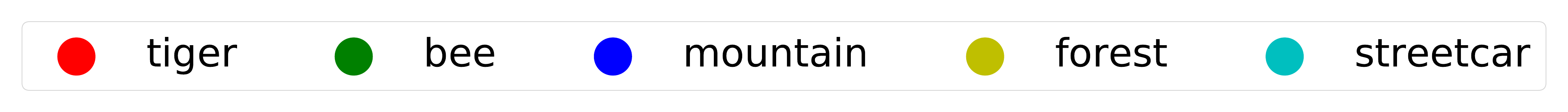}
    }
\caption{Embedding visualization after TSNE with our proposed method (left), and prototypical network (right).}
    \label{Tsne_compare}
\end{figure}

\noindent\textbf{TSNE visualization} We  visualize the embedding features of five categories randomly chosen from $C_{novel}$  (See Figure~\ref{Tsne_compare}). It could be seen that with our proposed method, categories that are similar in concepts tend to be closer to each other. For example, all the non-living things (mountain, forest, streetcar) are located in the top area while living things (bee, tiger) are placed bottom. Mountain and forest are the nearest two clusters since they are ``all outdoor scenes'' and their semantic distance is the closest among the five categories. On the other hand, the  prototypical network could successfully separate the five categories apart from each other, but they are located randomly in the graph without considering the semantic relationships.

\section{Conclusion}

Few-shot learning is typically under label-based supervision, which discards the semantic relationships and fails to make a  class association when there is no label matching between support and query set.  However, humans could easily identify the right association by considering a higher-level concept. Inspired by this, we present a psychometric testing based method that could capture images' high-level conception relationships to address the challenge. We evaluate our method on CIFAR-100 dataset. The results indicate that our method is capable of achieving higher semantic-based few-shot learning accuracy even with fewer annotating burdens than the baseline.



\bibliography{ref}
\end{document}